%% file: main.tex
\input{settings}
\input{acronym-setup}
\begin{document}

\maketitle

\begin{abstract}
Engineering models created in Model-Based Systems Engineering (MBSE) environments contain detailed information about system structure and behavior. 
However, they typically lack symbolic planning semantics such as preconditions, effects, and constraints related to resource availability and timing. 
This limits their ability to evaluate whether a given system variant can fulfill specific tasks and how efficiently it performs compared to alternatives.

To address this gap, this paper presents a model-driven method that enables the specification and automated generation of symbolic planning artifacts within SysML-based engineering models. 
A dedicated SysML profile introduces reusable stereotypes for core planning constructs. 
These are integrated into existing model structures and processed by an algorithm that generates a valid domain file and a corresponding problem file in  Planning Domain Definition Language (PDDL). 
In contrast to previous approaches that rely on manual transformations or external capability models, the method supports native integration and maintains consistency between engineering and planning artifacts. 

The applicability of the method is demonstrated through a case study from aircraft assembly. 
The example illustrates how existing engineering models are enriched with planning semantics and how the proposed workflow is applied to generate consistent planning artifacts from these models. 
The generated planning artifacts enable the validation of system variants through AI planning.

\end{abstract}

\section{Introduction}

Engineering automated production systems involves complex decisions that affect not only the physical structure of a system but also its operational feasibility and efficiency. 
\ac{mbse} has become a widely adopted approach to support this process by enabling consistent, tool-supported modeling of system structure and behavior~\cite{Henderson.2021}. 
Through integrated models, engineers can represent cross-disciplinary aspects of a system and maintain consistency throughout the development lifecycle~\cite{Schmidt.2020}.

\ac{sysml} provides a standardized notation for creating such models and supports the representation of structure, interfaces, and system functions~\cite{weilkiens2016sysmod}. 
While \ac{sysml} models capture rich technical detail, they typically do not include planning semantics such as action preconditions, resource constraints, or temporal dependencies. 
These aspects are essential for task-oriented planning and decision support.

As a result, engineers are often unable to use existing system models to answer critical planning questions.  Examples include whether a specific system variant can fulfill a given task, or how well it performs compared to alternative system variants. 
These questions are essential for validating functional feasibility and for evaluating key performance indicators such as cycle time, idle time, or reconfiguration effort~\cite{Törmanen.2017}. 

Symbolic \textit{AI Planning} provides formal mechanisms to generate action sequences that satisfy goal conditions under defined constraints~\cite{Ghallab.2016}. 
\ac{pddl} is the established standard for describing such planning problems~\cite{Haslum.2019}. 
However, manually creating \ac{pddl} descriptions is time-consuming and error-prone~\cite{Lindsay.2023, Sleath2024}. 
Translating engineering models into planning representations typically requires additional manual effort and introduces a disconnect between the system model and the planning logic~\cite{Kocher.2022}. 
While planning tools such as domain editors or syntax checkers exist~\cite{Strobel2020MyPDDL}, they are external to engineering workflows and require expertise not typically found among system engineers.

A central challenge is that system models and planning descriptions follow fundamentally different modeling logics. 
System models focus on structural and behavioral aspects of technical systems, emphasizing physical consistency and implementation detail. 
Planning descriptions, in contrast, rely on an abstract, symbolic representation of actions, including preconditions and effects, to evaluate goal satisfaction under given constraints. 
Translating between them requires not only syntactic conversion, but also a consistent way to represent planning semantics within the structure of engineering models.

To address this limitation, a structured workflow was developed to connect system and product models with planning semantics and to enable automated generation of \ac{pddl} domain and problem files from engineering artifacts~\cite{Nabizada2024b}. 
As part of this workflow, a dedicated \ac{sysml} profile was introduced to allow planning constructs such as types, predicates, and actions to be embedded directly into system models~\cite{Nabizada2025ETFA}. 
A transformation algorithm complements the approach by extracting the annotated model content and generating syntactically valid \ac{pddl} files~\cite{Nabizada2024c}.

This paper integrates the previously introduced workflow, \ac{sysml} profile, and transformation algorithm into a complete method for generating planning descriptions from engineering models. The focus lies on demonstrating the practical application of this approach in a case study from aircraft assembly, including the enrichment of engineering models with planning semantics and the automated generation of \ac{pddl} files.

The remainder of this paper is structured as follows. 
Section~\ref{sec:relatedwork} discusses related work on planning model generation and MBSE integration. 
Section~\ref{sec:workflow} outlines the four-phase workflow for embedding planning semantics into system and product models. 
Section~\ref{sec:pddlprofile} introduces the SysML profile used to represent planning constructs. 
Section~\ref{sec:algorithm} describes the transformation algorithm for generating PDDL files. 
Section~\ref{sec:application} presents a case study from aircraft assembly and evaluates the resulting plans. 
Finally, Section~\ref{sec:conclusion} summarizes the contributions and discusses directions for future research.

\section{Related Work}
\label{sec:relatedwork}

A range of approaches exists for generating planning descriptions from structured representations. 
These differ in modeling formalism, degree of automation, and the handling of planning semantics. 

\citet{Huckaby.2013} introduced a SysML-based taxonomy for assembly tasks that provides the basis for deriving PDDL actions manually. 
The approach supports the representation of system capabilities but does not automate the creation of planning files and is limited to predefined taxonomic structures. 

\citet{vieira2023transformation} proposed an ontology-based transformation of capability descriptions into PDDL by matching required and offered functions. 
While this enables automated generation of planning descriptions, it relies on a separate capability model and does not leverage existing engineering models.

\citet{Rimani.2021} propose a conceptual mapping between SysML-based functional architectures and the \ac{hddl}. 
The approach enables the structuring of hierarchical planning domains from engineering models but remains largely manual. 
It lacks a dedicated profile, formal validation mechanisms, and support for traceability between planning artifacts and system models. 

\citet{wally2019flexible} present a model-driven approach for generating PDDL representations from ISA-95-compliant manufacturing system models. 
The method defines metamodel-level mappings and supports automated generation of domain and problem files. 
While effective in industrial automation contexts, it is limited to ISA-95-compliant models, making it unsuitable for integration with general-purpose modeling frameworks such as SysML.

\citet{Konidaris.2018} present a data-driven approach for learning symbolic AI planning models from sensorimotor trajectories. 
The method derives abstract state transitions and operator models by analyzing the effects of high-level robotic skills in continuous environments. 
The resulting models are tailored to specific tasks and environments and are not parameterized or linked to reusable engineering artifacts. 
As such, they are not suitable for integration with structured system modeling environments where consistency, traceability, and reuse are required. 

\citet{Stoev.2023} present a tool that automatically extracts symbolic planning models in PDDL from instructional texts. 
While effective in knowledge extraction from unstructured sources, the approach is not integrated into model-based engineering workflows and does not leverage formal system representations. 

Recent work has explored the use of large language models (LLMs) for generating symbolic AI planning models from natural language input. 
\citet{Tantakoun2024LLMModelers} provide a comprehensive survey and taxonomy of this emerging line of research, framing LLMs as modelers rather than solvers. They distinguish between model generation, model editing, and model benchmarking, and highlight the potential of LLMs to create domain and problem descriptions in PDDL from textual input.
These methods demonstrate strong syntactic capabilities and can generate plausible planning structures without dedicated modeling tools. However, they often suffer from limited semantic accuracy and inconsistencies in domain logic. While hybrid approaches combining LLM output with validation tools can mitigate some issues, the generated models typically lack integration with engineering data sources and are not traceable to structured system descriptions.

In contrast, the approach presented in this paper builds on formally defined system and product models created in MBSE environments. 
Rather than inferring planning knowledge from text, it uses stereotype annotations to embed planning constructs directly into SysML artifacts. 
This ensures consistency with system architecture and enables traceable and automated generation of domain and problem files from verified engineering models.

\section{Workflow for Automated Generation of PDDL Descriptions}
\label{sec:workflow}

In symbolic planning, a plan consists of a sequence of actions that transform an initial state into a goal state, based on a formally defined planning domain~\cite{Ghallab.2016}. 
Each action has parameters, preconditions, and effects, and may depend on resource availability, spatial relations, or system constraints such as energy or tool compatibility. 
To generate a valid plan, the domain must define object types, predicates, and actions, while the problem instance specifies the concrete objects involved, the initial conditions, and the goal~\cite{Haslum.2019}. 
This structure imposes specific requirements on the modeling of technical systems: symbolic relations between system elements must be made explicit, and behavioral aspects must be translated into symbolic actions with parameters, preconditions, and effects. 

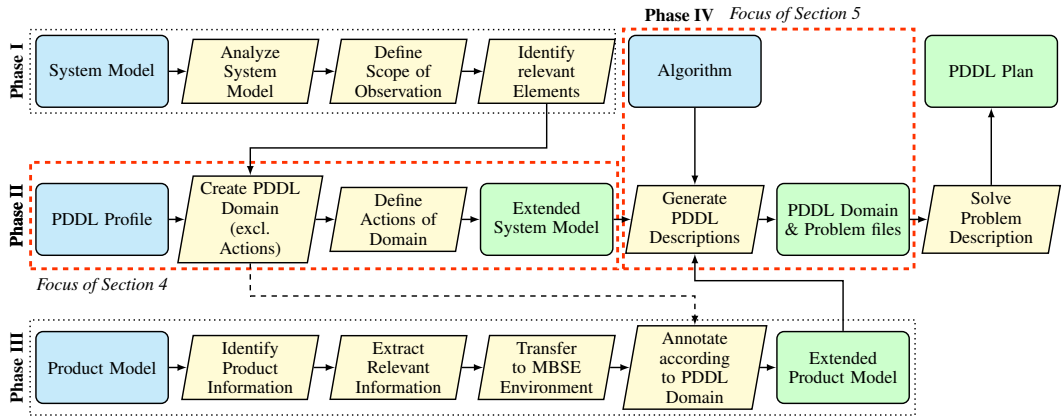
\begin{figure*}[h!]
    \centering
    \resizebox{0.8\linewidth}{!}{
\input{figures/workflow.tex}
    }
    \caption{Workflow model for automated generation of PDDL descriptions (adapted from~\citet{Nabizada2024b}).}
    \label{fig:workflow_model}
\end{figure*}

Integrating planning capabilities into engineering models enables consistent evaluation of system variants and task feasibility. 
In practice, however, system and product data are often maintained in separate tools, making it difficult to generate and maintain consistent planning artifacts. 
To address this, a structured workflow was developed that connects SysML-based system models with product data and formal planning constructs. 
The workflow supports the automated generation of a \ac{pddl} domain and a \ac{pddl} problem file and was introduced in earlier work~\cite{Nabizada2024b}. 
An overview is shown in Figure~\ref{fig:workflow_model}.

\definecolor{HighlightPhase}{RGB}{255,50,0} 

\textbf{Phase I} addresses the challenge of working with existing system models that were not originally created with planning tasks in mind. 
These models often contain heterogeneous information, including structural, behavioral, and functional aspects of technical systems~\cite{Henderson.2021, Schmidt.2020}. 
Users applying the workflow are typically not involved in the creation of the model and must therefore analyze its content to understand the system architecture and the terminology used~\cite{Nabizada2024b}.

Since such models may describe complex and large-scale systems, for example entire production facilities~\cite{Törmanen.2017}, it is necessary to restrict the scope of analysis. 
In many cases, the planning problem concerns only a specific machine, an individual production cell, or a functional module. 
A scoping step is therefore required to isolate the parts of the model that are relevant for planning and to exclude unrelated information. 
Within this defined scope, relevant components, resources, and their relationships are identified. 
This step ensures that the subsequent enrichment of the model focuses only on information needed for the planning domain~\cite{Nabizada2024b}. 
To gain this understanding, engineers must review the model structure, analyze relevant block and activity diagrams, and identify consistent terminology and modeling patterns that describe system elements and their behavior. 

In \textbf{Phase II}, the system model is extended with planning semantics using a dedicated \ac{sysml} profile~\cite{Nabizada2025ETFA}. 
This profile introduces stereotypes for core \ac{pddl} constructs and allows the integration of domain knowledge directly into the model. 
The enrichment begins with the specification of object types and predicates, which define structural elements and relationships relevant for planning.

Based on these definitions, actions are defined that describe system behavior in terms of parameters, preconditions, and effects. 
All planning elements are modeled consistently within the system architecture and linked to existing structures where applicable. 
The result is an extended system model that contains the complete domain information required for generating a valid \ac{pddl} domain file. 
The structure and function of the SysML profile used in this phase are described in more detail in Section~\ref{sec:pddlprofile}. 

In \textbf{Phase III}, information from the product model is incorporated to define the specific planning task. 
While the system model provides the available structure and capabilities, the product model contains instance-level data such as the positions and types of components required for problem definition~\cite{beers2024towards}. 
Relevant information is extracted from the product model and transferred into the MBSE environment, where it is annotated using the planning concepts previously defined in the system model. 
By referencing the same types and predicates, consistency between domain and problem description is ensured. 
The result is an extended product model, integrated into the MBSE environment, that serves as the basis for generating the \ac{pddl} problem file. 

In \textbf{Phase IV}, the information embedded in the system and product models is used to automatically generate the corresponding \ac{pddl} domain and problem file.
Since all relevant planning elements have been defined in the previous steps, this generation can be carried out without additional manual effort. 
A transformation algorithm processes the annotated model content and converts it into syntactically correct PDDL descriptions using a template-based approach. 
The details of this algorithm are described in Section~\ref{sec:algorithm}.

The resulting domain file contains all types, predicates, and actions as specified in the system model, while the problem file reflects the initial state and goal conditions derived from the product data. 
These files can be passed to a standard PDDL solver to compute a valid plan. 
Because the planning artifacts are derived directly from the models, changes to the system or product can be propagated automatically, ensuring that the planning logic remains aligned with the current system variant.

\section{SysML Profile for PDDL Integration}
\label{sec:pddlprofile}
Standard SysML\footnote{For a detailed introduction to SysML concepts 
including stereotypes, metaclasses, and UML-based metamodeling, 
see~\cite{Friedenthal.2014}.}  does not provide native constructs for expressing symbolic planning semantics such as typed action parameters, logical preconditions, or effects. 
To address this, a dedicated \ac{sysml} profile was developed and introduced in~\citet{Nabizada2025ETFA}. 
It is based on the \ac{bnf} specification of PDDL~3.1~\cite{Kovacs.2011} and enables core planning concepts to be represented as stereotypes within \ac{sysml} models. 
The profile supports model-based specification of planning descriptions and enables the consistent generation of \ac{pddl} files from annotated engineering models. 
An overview of the mapping between \ac{pddl} constructs and \ac{sysml} elements is provided in Table~\ref{tab:pddl_to_sysml}.

\begin{table}[h]
\centering
\scriptsize
\begin{tabular}{>{\raggedright\arraybackslash}p{2.0cm} >{\raggedright\arraybackslash}p{2.3cm} >{\raggedright\arraybackslash}p{2.4cm}}
\toprule
\textbf{PDDL Concept} & \textbf{Applied Stereotype} & \textbf{Extended SysML Metaclass} \\

\midrule
Domain Definition & \texttt{<<PDDL\_Domain>>} & \texttt{Model}, \texttt{Package} \\
Object Type & \texttt{<<PDDL\_Type>>} & \texttt{Class} \\
Logical Condition & \texttt{<<PDDL\_Predicate>>} & \texttt{ObjectFlow}, \newline \texttt{ControlFlow} \\
Numeric Expression & \texttt{<<PDDL\_Function>>} & \texttt{ObjectFlow}, \newline \texttt{ControlFlow} \\
Action Specification & \texttt{<<PDDL\_Action>>} & \texttt{CallBehaviorAction} \\

\bottomrule
\end{tabular}
\caption{Mapping of PDDL Concepts to SysML Elements~\cite{Nabizada2025ETFA}}
\label{tab:pddl_to_sysml}
\end{table}

Each planning construct defined in the PDDL specification is represented in the profile by a dedicated stereotype. 
These stereotypes extend selected metaclasses from the SysML or UML metamodel and serve to embed planning-specific semantics into existing modeling constructs. 

The stereotype \stereotyp{Domain} is applied to elements of type \sysml{Model} or \sysml{Package}. 
It provides the structural container for a planning domain and groups all relevant model elements, including types, predicates, and actions, under a consistent planning context. 
This approach allows domain definitions to be integrated into larger system models while remaining encapsulated.

Object types are defined using \stereotyp{Type}, which extends the metaclass \sysml{Class}. 
These types are used to categorize domain objects and to define the types of symbolic parameters used in planning actions. 
Such parameters represent abstract references to objects manipulated by an action, for example the part to assemble or the tool to use. 
Inheritance relations between types can be modeled using generalization mechanisms available in \ac{uml} and \ac{sysml}.

Logical conditions and numerical values are described using \stereotyp{Predicate} and \stereotyp{Function}. 
These stereotypes are applied to \sysml{ObjectFlow} and \sysml{ControlFlow} elements in activity diagrams. 
Predicates represent logical conditions relevant to the planning logic, such as whether a resource is available. 
Functions allow the use of numeric data in planning, for example cost, duration, or distance.

Actions are modeled using \stereotyp{Action}, which extends the metaclass \sysml{CallBehaviorAction}. 
This integration allows planning behavior to be described within standard \ac{sysml} activity diagrams. 
Action parameters are defined based on previously declared types, while preconditions and effects reference predicates and functions. 
The use of \sysml{CallBehaviorAction} aligns planning actions with commonly used behavioral modeling elements~\cite{beers2024sysml}. 

To clarify the role of the SysML profile within Phase~II of the workflow (see Figure~\ref{fig:workflow_model}), Listing~\ref{lst:assemble_part} shows an example of a PDDL action that can be automatically generated from a suitably annotated system model. In a typical system model, elements such as \textit{tool} and \textit{part} are already defined as structural types, and behavioral steps like \textit{assemble-part} are modeled as actions with incoming and outgoing flows. During Phase~II, these elements are enriched with planning semantics by applying the profile's stereotypes. This annotation enables the extraction of planning-relevant logic and the generation of consistent PDDL actions.

\begin{listing}[h]
\caption{Example PDDL action: \element{assemble-part}}
\label{lst:assemble_part}
\begin{lstlisting}[language=PDDL, 
backgroundcolor = \color{light-gray},
basicstyle=\ttfamily\footnotesize]
(:action assemble-part
 :parameters (?p - part ?t - tool)
 :precondition (available ?t)
 :effect (assembled ?p))
\end{lstlisting}
\end{listing}

In this example, the object types \texttt{part} and \texttt{tool} correspond to classes stereotyped as \stereotyp{Type}. 
The action node representing the assembly process is annotated as \stereotyp{Action}, while the flows that express the required and resulting state are marked with \stereotyp{Predicate}. 
These annotations define the structure and semantics needed to derive the PDDL action shown above through model transformation. 
Since the correctness of the generated planning artifacts relies on the internal consistency of the annotated model, syntactic and structural validation is required, particularly in large or evolving system descriptions.

To support this, the profile integrates formal validation mechanisms based on the \textit{Object Constraint Language}~(OCL). 
Constraints are defined for each stereotype and can be evaluated during model editing. 
This allows for early detection of modeling errors and promotes consistent use of planning semantics across the system model.

Listing~\ref{lst:ocl_typenames} shows a constraint applied to \stereotyp{Domain} elements. 
It checks that each contained \stereotyp{Type} element has a unique name. This prevents ambiguous type references and avoids errors during domain interpretation~\citep{Sleath2024}.

\begin{listing}[h]
\caption{OCL constraint for enforcing unique type names within a domain~\cite{Nabizada2025ETFA}}
\label{lst:ocl_typenames}
\begin{lstlisting}[language=OCL, 
backgroundcolor = \color{light-gray},
basicstyle=\ttfamily\footnotesize] 
context PDDL_Domain inv UniqueTypeNames: 
  self.ownedElement 
    ->select(e | e.oclIsKindOf(PDDL_Type)) 
    ->isUnique(t | t.name) 
\end{lstlisting}
\end{listing}

The validation logic is integrated directly into the profile and can be processed by tools that support OCL evaluation. In complex planning domains, modeling errors such as redundancies or inconsistencies can occur even in well-structured models, particularly when tool support for validation and feedback is limited~\citep{Shah2013}.

The profile is published as an open resource\footnote{\url{https://github.com/hsu-aut/SysML-Profile-PDDL}} and can be applied in modeling environments to support planning integration. 
It provides a general mechanism for embedding planning constructs into system models and supports both domain and problem description, depending on the available model content.

\section{Algorithm for PDDL Domain File Generation}
\label{sec:algorithm}
The transformation algorithm applied in Phase~IV of the workflow (see Figure~\ref{fig:workflow_model}) automatically derives a complete \ac{pddl} domain description from a SysML system model enriched with planning semantics. 
Its input consists of model elements annotated with stereotypes defined in Section~\ref{sec:pddlprofile}, including types, predicates, functions, and actions. 
The result is a syntactically valid and semantically consistent domain file that can be submitted to any standard \ac{pddl} solver.

The overall process follows a sequential logic that extracts, interprets, and formats the planning-relevant elements. 
Its core workflow is illustrated as pseudocode in Algorithm~\ref{alg:pddl-generation}. 
The procedure begins by initializing the domain context and continues by collecting all model elements annotated with the relevant stereotypes. 
Each planning concept is then translated into its corresponding structure in the \ac{pddl} domain description.

\renewcommand{\algorithmicrequire}{\textbf{Input:}}
\renewcommand{\algorithmicensure}{\textbf{Output:}}
\algnewcommand\algorithmicforeach{\textbf{for each}}
\algdef{S}[FOR]{ForEach}[1]{\algorithmicforeach\ #1\ \algorithmicdo}
\begin{algorithm}[h]
    \footnotesize
    \caption{Generation of the PDDL domain file~\cite{Nabizada2024c}}
    \label{alg:pddl-generation}  
    \begin{algorithmic}[1]
    \Require{SysML-based system model \textit{sm}}
    \Ensure{PDDL domain file \textit{df}}
    \Function{CreatePDDLDomain}{\textit{sm}}
        \State \textit{df} $\leftarrow$ \Call{InitializePDDLDomain}{\textit{sm.PDDL\_Domain}}
        \State \textit{T} $\leftarrow$ \Call{ExtractTypes}{\textit{sm.PDDL\_Type}}
        \ForEach{\textit{type} in \textit{T}}
            \State \Call{ProcessTypeToPDDL}{\textit{df}, \textit{type}}
        \EndFor
        \State \textit{P} $\leftarrow$ \Call{ExtractPredicates}{\textit{sm.PDDL\_Predicate}}
        \ForEach{\textit{predicate} in \textit{P}}
            \State \Call{ProcessPredicateToPDDL}{\textit{df}, \textit{predicate}}
        \EndFor
        \State \textit{A} $\leftarrow$ \Call{ExtractActions}{\textit{sm.PDDL\_Action}}
        \ForEach{\textit{action} in \textit{A}}
            \State \Call{ProcessActionToPDDL}{\textit{df}, \textit{action}}
        \EndFor
        \State \Return \textit{df}
    \EndFunction
    
    \end{algorithmic}
    
\end{algorithm}

The algorithm starts by initializing the PDDL domain structure based on the model element annotated with \stereotyp{Domain} (line~2), before extracting planning-relevant elements using stereotype-based model queries. 
All classes annotated with \stereotyp{Type} are processed recursively to reflect inheritance hierarchies~(lines~3–6). 
Logical predicates and numeric functions are extracted from control and object flows that are annotated with \stereotyp{Predicate} or \stereotyp{Function}~(lines~7–10). 
For each element, variable names and types are resolved and checked for consistency. 
Missing definitions are reported to the user to prevent the generation of an invalid domain file.

Behavioral actions are derived from elements annotated with \stereotyp{Action}~(lines~11–14). 
These actions are modeled as \texttt{CallBehaviorAction} nodes and define parameters, preconditions, and effects. 
The algorithm analyzes incoming and outgoing flows to determine the associated logical conditions. 
The algorithm examines the flows connected to each action to identify its preconditions and effects.
It supports both positive and negative predicates, as well as combinations of multiple conditions using logical conjunctions like \texttt{and}. 
Numerical effects, such as cost increments, are expressed using PDDL functions that update numeric values during planning. 
The resulting \ac{pddl} action definitions reflect the logic embedded in the system model and remain directly traceable to their origin.

The transformation of annotated SysML models into PDDL descriptions is implemented using the \textit{Apache Velocity Engine}. 
Templates written in the \textit{Velocity Template Language} (VTL)\footnote{Available at \url{https://github.com/hsu-aut/VTL-PDDL_Domain}} define the syntactic structure of the output and inject model content into predefined sections. 
Planning-relevant elements such as types, predicates, functions, and actions are identified through stereotype-specific model queries and converted into PDDL syntax.

Listing~\ref{lst:pddl_predicates} shows an excerpt of the generated \texttt{:predicates} section, based on stereotyped flows within the system model.

\begin{listing}[h]
\caption{Excerpt from a generated PDDL predicate section}
\label{lst:pddl_predicates}
{\footnotesize
\begin{lstlisting}[backgroundcolor = \color{light-gray},
  language=PDDL,
  basicstyle=\ttfamily\footnotesize]
(:predicates
    (is-available ?r - Resource)
    (connected ?a - Location ?b - Location)
)
\end{lstlisting}
}
\end{listing}

Each parameter and type is extracted from the annotated model and checked for completeness.
If required names or references are missing, the transformation inserts explicit error messages into the generated PDDL code.
These embedded checks help identify modeling issues early and prevent incomplete or inconsistent planning constructs from being generated without notice. 
While not a substitute for formal validation, these checks complement the OCL constraints defined in the SysML profile by covering additional technical aspects required for code generation. 
They provide immediate feedback during transformation and help ensure the completeness of the generated PDDL files.

A key advantage of the transformation approach is its alignment with the modeling perspective of system engineers.
All planning constructs are embedded directly into the SysML model, avoiding the need to switch between external tools or learn additional planning languages.
This integration ensures that the system model remains the single source of truth, reducing inconsistencies and supporting traceable generation of planning artifacts.

The use of reusable templates enables flexible adaptation to domain-specific constructs, solver-specific syntax, or alternative planning dialects.
Because the transformation is fully embedded in the modeling environment, it supports repeatable workflows and consistent integration of planning logic across engineering projects.

\section{Application in Aircraft Assembly}
\label{sec:application}

To evaluate the approach in a realistic engineering context, a case study from aircraft structure assembly was conducted. 
The use case forms part of the iMOD research project~\cite{Gehlhoff2022iMOD} and addresses the automated planning of collar screwing operations inside the fuselage. 
This process involves a UR10 robotic arm equipped with interchangeable end effectors, each capable of processing a specific collar type. 
Rivets are fasteners used to join structural components, while collars are threaded sleeves that are screwed onto the rivet shafts from the inside to secure the connection. 
The robot operates in a constrained environment within the fuselage and must reach multiple rivet positions while performing tool changes as required by the collar types. 
The objective is to generate an optimized plan that minimizes cycle time and reconfiguration effort while ensuring all rivets are processed correctly. 



The underlying system model used in this work was developed following the structured modeling workflow presented in~\cite{beers2023mbse}. 
It provides a detailed representation of the \textit{Collar Screwing System}~(CSS), including stakeholder requirements, functional decomposition, and technical subsystem structures. 
The model was created and extended using \ac{msosa}, an MBSE environment with SysML support. 
It served as the basis for applying the model-based workflow described in Section~\ref{sec:workflow}. 

\subsection{Phase I}

In \textbf{Phase I}, the existing SysML model of the CSS~\cite{beers2023mbse} was analyzed to identify model content relevant for the planning task. 
Since the system model covers multiple engineering aspects, including control logic, vision systems, and user interaction, it was necessary to restrict the observation scope to elements directly related to the physical execution of screwing operations.

This included, for example:
\begin{itemize}
    \item the UR10 robotic manipulator as execution unit,
    \item interchangeable screwing tools as subsystems,
    \item the structural description of rivets as domain-relevant types,
    \item and actions for motion and assembly behavior.
\end{itemize}

While instance-level rivet positions are part of the product model considered in Phase~III, the abstract description of rivet types and their functional role is embedded in the system model and relevant for domain specification. 

Based on this scoping, a subset of the model was defined as the foundation for introducing planning semantics in Phase~II. 
The selected model elements reflect the key resources, object types, and behavioral steps required for defining symbolic actions and constraints. 
This focused selection ensures that the planning logic remains aligned with the system architecture while avoiding unnecessary complexity by excluding subsystems that are out of scope.

\subsection{Phase II}
In \textbf{Phase II}, the relevant parts of the system model were enriched with planning semantics using the SysML profile introduced in Section~\ref{sec:pddlprofile}. 
Stereotypes were applied to define PDDL types (e.g., \element{Rivet}, \element{ScrewingTool}), predicates (e.g., \element{CollarScrewed}, \element{EnergySupply}), and functions (e.g., \element{RivetDistanceInformation}).

\begin{figure}[h]
\centering
\includegraphics[width=0.55\linewidth]{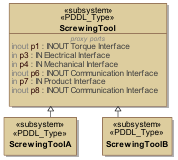}
\caption{PDDL type annotation for screwing tool variants}
\label{fig:pddl_types}
\end{figure}

Figure~\ref{fig:pddl_types} shows how structural components of the system, such as end effectors, were annotated with the stereotype \stereotyp{Type} to define object types used in planning. 
The subsystems \element{ScrewingToolA} and \element{ScrewingToolB} are modeled as tool variants and form the basis for typing action parameters in the generated \ac{pddl} domain. 
Interfaces such as mechanical, electrical, and communication ports remain part of the system model but are not explicitly represented in the planning description. 

\begin{figure*}[h]
\centering
\includegraphics[width=0.75\linewidth]{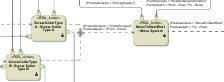}
\caption{Excerpt from the enriched system model of the collar screwing system~\cite{Nabizada2025ETFA}}
\label{fig:systemmodel_excerpt}
\end{figure*}

Based on the defined types and predicates, planning-relevant system behavior was formalized through annotated actions.

Figure~\ref{fig:systemmodel_excerpt} shows an excerpt from the activity diagram in the MBSE environment. The actions \element{ScrewCollarTypeA}, \element{ScrewCollarTypeB}, and \element{MoveToNextRivet} are annotated with the stereotype \stereotyp{Action} and define planning-relevant logic in terms of parameters, preconditions, and effects. 
Logical conditions such as energy availability or the completion of screwing operations are modeled using stereotyped predicates, while numerical values such as travel cost are defined via functions. 
These annotations allow the behavioral logic of the system to be translated directly into structured \ac{pddl} actions. 

The result is an extended system model in which planning constructs are embedded alongside the existing structural and behavioral descriptions. 
This model provides a consistent and traceable basis for generating the \ac{pddl} domain file during Phase~IV of the workflow.

\subsection{Phase III}

In \textbf{Phase III}, instance-level product data was incorporated to define the specific planning problem. 
While the system model describes general capabilities and structural elements of the CSS, the product model provides configuration-specific information required to instantiate a concrete planning task. 
In this case, the product model was maintained in \textit{Dassault Systèmes 3DExperience} and contained the geometry, type, and position of rivets to be processed in a particular aircraft section.

Relevant product data was extracted from the product structure model and transferred into \ac{msosa}. 
This information includes the unique identifiers and spatial coordinates of the rivets as well as their assigned collar types. 
To align the product data with the previously defined planning domain, each instance was annotated using the same \ac{pddl} types and predicates introduced in Phase~II. 
For example, each rivet instance was typed as \element{Rivet} and linked to its location and assembly status via appropriate predicates such as \element{CollarScrewed}. 

This integration results in an extended product model that complements the enriched system model. 
It defines the initial state of the planning problem and specifies the goal conditions, such as all collars being screwed at their designated positions. 
By referencing domain-level types and predicates, consistency between domain and problem descriptions is ensured. 

\subsection{Phase IV}

In \textbf{Phase IV}, the enriched system and product models were used to automatically generate the corresponding \ac{pddl} domain and problem file. 
Since all relevant planning semantics were defined through model annotations in the previous phases, no additional manual modeling steps were required. 
Instead, the transformation algorithm (see Section~\ref{sec:algorithm}) was applied that systematically extracts model content based on the applied stereotypes and converts it into syntactically correct PDDL code.

In the applied modeling environment, the transformation is executed using the integrated \textit{Report Wizard}. 
It provides direct access to the annotated model and invokes the \textit{Apache Velocity Engine} to process VTL-based templates. These templates retrieve stereotype-specific content from the model and assemble it into structured domain and problem file. The transformation is embedded in the modeling tool and does not require external scripts or additional software components. This setup enables reproducible file generation directly within the MBSE environment.

Figure~\ref{fig:pddl_types} and Figure~\ref{fig:systemmodel_excerpt} shows how object types and actions were defined within the SysML model using the profile introduced in Section~\ref{sec:pddlprofile}. 
The corresponding PDDL code fragments are generated automatically and reflect these structures in symbolic form.

Listing~\ref{lst:pddl_types} shows an excerpt of the generated \texttt{:types} section, capturing the type hierarchy defined in the system model through stereotyped class elements.

\begin{listing}[h]
\caption{Excerpt from the generated PDDL types}
\label{lst:pddl_types}
\begin{lstlisting}[language=PDDL, 
backgroundcolor = \color{light-gray},
basicstyle=\ttfamily\footnotesize]
(:types
 ; ... (omitted types)
  ScrewingTool - CSS
  ScrewingToolA ScrewingToolB - ScrewingTool
)
\end{lstlisting}
\end{listing}

The action logic is derived from the behavioral annotations within the activity diagram shown in Figure~\ref{fig:systemmodel_excerpt}. 
The model element \element{MoveToNextRivet} is annotated with the stereotype \stereotyp{Action} and includes planning-relevant constructs such as preconditions (\element{CollarScrewed}, \element{EnergySupply}) and numerical effects (\element{RivetDistanceInformation}). 
These annotations are automatically translated into structured PDDL syntax.

Listing~\ref{lst:pddl_action} illustrates the generated representation of this action, including typed parameters, logical conditions, and a cost-based effect modeled using a numeric function.
In total, the generated domain contained 27 types, 21 predicates, 
3 functions, and 12 actions.

\begin{listing}[h]
 \caption{Generated PDDL action for \element{MoveToNextRivet}}
 \label{lst:pddl_action} 
 
\begin{lstlisting}[language=PDDL, 
backgroundcolor = \color{light-gray},
basicstyle=\ttfamily\footnotesize]
(:action MoveToNextRivet
 :parameters (?from - Rivet ?to - Rivet)
 :precondition 
   (and 
     (CollarScrewed ?from)
     (EnergySupply)
   )
 :effect 
   (and  
     (MovedToNextRivet ?to) 
     (increase 
       (total-cost) 
       (RivetDistanceInformation ?from ?to)) 
   )
)
\end{lstlisting}
\end{listing}

\subsection{Application Results}
\label{sec:evaluation}

Based on the automatically generated \ac{pddl} domain and problem files, the analysis tool \textit{VAL}~\cite{Howey2004VAL} was used to check their syntactic and semantic correctness. 
\textit{VAL} reported no errors or warnings, confirming that all types, predicates, and actions were consistently defined and that no undefined symbols or unreachable actions were present.

After validation, the \textit{Delfi} planner~\cite{Katz.2018} was used to compute a valid plan for the evaluated system variant. 
\textit{Delfi} is a heuristic-based planner for cost-optimal planning and was accessed via \textit{Planutils}~\cite{Muise.2022}, which provides a standardized interface for invoking a wide range of planning tools. 
The plan was validated using \textit{VAL}~\cite{Howey2004VAL}, confirming that it satisfies the goal conditions and is consistent with the domain specification. 

The resulting plan defines the sequence of screwing and movement operations required to assemble all rivets, taking into account tool constraints and spatial dependencies. 
In the evaluated scenario, two different rivet types were placed within the robot's workspace, each requiring a dedicated end effector. 
The planner minimized overall process time by considering travel distances between rivet positions and the overhead of tool changes. 
As a result, the plan grouped operations by tool type to avoid unnecessary reconfigurations, thereby reducing motion overhead and idle time. 
This execution strategy reflects a potentially optimized cycle time and can be used as a reference for evaluating alternative system variants.

The case study confirms the practical applicability of the proposed approach. 
It demonstrates how the integration of planning semantics into engineering models enables the automated generation of consistent planning artifacts and supports traceable variant analysis. 
Although applied here in the context of aircraft assembly, the method is domain-independent and applicable to other engineering workflows involving symbolic reasoning and decision support.

However, the results also highlight certain limitations that merit further investigation. 
Although the automated transformation reduces the need for manual effort, the quality of the resulting plans still depends on the precision and completeness of the model, as well as the annotation process.
In particular, the usability of the modeling environment and the clarity of the profile elements can affect modeling speed and accuracy. 
Future improvements should focus on simplifying the modeling process itself, for example through guided annotation workflows or semi-automated suggestions for stereotype application.

\section{Conclusion and Future Work}
\label{sec:conclusion}

This paper presents a model-driven approach for generating planning descriptions in \ac{pddl} from engineering models. 
The method combines a structured workflow, a dedicated SysML profile for embedding planning semantics, and a transformation algorithm for generating consistent PDDL files. 

The approach was applied in a case study from aircraft structure assembly, demonstrating how system behavior and product data can be enriched with symbolic planning constructs and used to automatically derive planning artifacts. 
The resulting domain file and corresponding problem file reflect the modeled system architecture and enable symbolic reasoning on configuration feasibility and task fulfillment. 
This confirms the practical relevance of the method for model-based variant evaluation and task-oriented system planning.

A key benefit lies in the traceable integration of planning semantics into engineering models. 
The approach supports the repeated generation of planning artifacts when system or product configurations change, thereby reducing manual effort and enabling consistent scenario analysis. 
The reuse of annotated model elements across variants further enhances modeling efficiency and planning reliability.

Future work will focus on lowering the entry barrier for planning integration. 
First, an interactive assistant is envisioned to support the annotation of planning constructs. 
Such a tool could guide users in applying stereotypes, validate structural consistency, and provide context-aware suggestions. 
Techniques based on rule inference or LLM-based prompting may assist in semi-automated annotation.

Second, further automation is needed in linking engineering data to planning tasks. 
In particular, the extraction and mapping of instance-level product information remains partially manual. 
Establishing an interface between tools such as \textit{3DExperience} and \ac{msosa} could enable direct reuse of product data within the modeling environment. 
This would reduce manual effort and support consistent integration of planning information across tool boundaries.

Additionally, compatibility with the \textit{Unified Planning Library}~\cite{Micheli2025UnifiedPlanning}, a Python-based API for modeling planning problems and interfacing with a wide range of planning engines, will be investigated to enable integration with standardized planning tools and workflows. 
Migration to SysML v2~\cite{ObjectManagementGroup.2024} is also planned to ensure continued applicability as modeling languages and tool support evolve.

\section*{Acknowledgments}
This research paper [project iMOD] is funded by dtec.bw– Digitalization and Technology Research Center of the Bundeswehr. dtec.bw is funded by the European Union – NextGenerationEU.

\renewcommand*{\bibfont}{\footnotesize}
\bibliography{references}

\end{document}

%% file: settings.tex
\documentclass[letterpaper]{article} 
\usepackage{aaai25}  
\usepackage{times}  
\usepackage{helvet}  
\usepackage{courier}  
\usepackage[hyphens]{url}  
\usepackage{graphicx} 
\usepackage{natbib}  
\usepackage{caption} 
\usepackage{algorithm}
\nocopyright

\usepackage{booktabs}
\usepackage{array}
\usepackage{tabularx} 
\usepackage{xcolor}
\usepackage{algpseudocode}

\usepackage{placeins}

\newcommand{\sysml}[1]{\texttt{#1}}
\newcommand{\stereotyp}[1]{\texttt{\textless\textless PDDL\_#1\textgreater\textgreater}}
\newcommand{\element}[1]{\textit{#1}}

\usepackage{tikz}
\usetikzlibrary{shapes, arrows.meta, positioning, fit}
\usetikzlibrary{decorations.pathreplacing}

\tikzset{
	>=Latex,
	line/.style={draw,->},
	anode/.style={rectangle,draw,
		align=center,rounded corners,minimum height=4em,font=\strut},
	bnode/.style={anode,fill=white, font=\strut},
	cnode/.style={anode, fill=cyan!20, font=\strut},
treenode/.style = {circle,
	draw=black,thick, fill=white, align=center, minimum size=1cm},
root/.style     = {treenode, font=\footnotesize},
env/.style      = {treenode, font=\footnotesize}, 
dummy/.style    = {circle,draw}
}
\tikzstyle{decision} = [diamond, draw, fill=yellow!20, 
    text width=6em, text badly centered, node distance=3cm, inner sep=0pt, line width=0.8pt, minimum height=9em,
    minimum width=9em]
\tikzstyle{block} = [rectangle, draw, fill=cyan!20, 
    text width=6.7em, text centered, rounded corners, minimum height=4em, line width=0.8pt]
\tikzstyle{line} = [draw, -latex, line width=0.8pt]
\tikzstyle{dashline} = [draw, -latex,dashed, line width=0.8pt]
\tikzstyle{helpline} = [draw, line width=0.8pt]
\tikzstyle{method} =  [trapezium, trapezium left angle=80, trapezium right angle=-80,text centered,text width = 2cm,minimum height=1cm, minimum width=2cm, draw=black, fill=yellow!20, line width=0.8pt]
\tikzstyle{exArtifact} =[rectangle, draw, fill=cyan!20, 
    text width=6.7em, text centered, rounded corners, minimum height=4em, line width=0.8pt]
\tikzstyle{genArtifact} =[rectangle, draw, fill=green!20, 
    text width=6.7em, text centered, rounded corners, minimum height=4em, line width=0.8pt]

\usepackage{newfloat}
\usepackage{listings}
\DeclareCaptionStyle{ruled}{labelfont=normalfont,labelsep=colon,strut=off} 
\floatstyle{ruled}
\newfloat{listing}{tb}{lst}{}
\floatname{listing}{Listing}

\input{listings-setup}
\title{Bridging Engineering and AI Planning through Model-Based Knowledge Transformation for the Validation of Automated Production System Variants}
\author{
    Hamied Nabizada\textsuperscript{\rm 1,*}, Lasse Beers\textsuperscript{\rm 1}, Alain Chahine\textsuperscript{\rm 2}, Felix Gehlhoff\textsuperscript{\rm 1}, \\Oliver Niggemann\textsuperscript{\rm 1}, Alexander Fay\textsuperscript{\rm 3}\\
}
\affiliations{
    \textsuperscript{\rm 1}Institute of Automation Technology, Helmut-Schmidt-University Hamburg, Hamburg, Germany\\
    \textsuperscript{\rm 2}Airbus Operations GmbH, Hamburg, Germany \\
    \textsuperscript{\rm 3}Chair of Automation, Ruhr University Bochum, Bochum, Germany\\
    
}

\usepackage{bibentry}

%% file: listings-setup.tex
\definecolor{light-gray}{gray}{0.95}
\lstdefinelanguage{PDDL}
{
  sensitive=false,    
  morecomment=[l]{;}, 
  alsoletter={:,-},   
  morekeywords={
    define,domain,problem,not,and,or,when,forall,exists,either,
    :domain,:requirements,:types,:objects,:constants,
    :predicates,:action,:parameters,:precondition,:effect,
    :fluents,:primary-effect,:side-effect,:init,:goal,
    :strips,:adl,:equality,:typing,:conditional-effects,
    :negative-preconditions,:disjunctive-preconditions,
    :existential-preconditions,:universal-preconditions,:quantified-preconditions,
    :functions,assign,increase,decrease,scale-up,scale-down,
    :metric,minimize,maximize,
    :durative-actions,:duration-inequalities,:continuous-effects,
    :durative-action,:duration,:condition
  }
}
\lstdefinelanguage{OCL}{
    morekeywords={context, inv, self, and, not, matches},
    sensitive=true,
    basicstyle=\footnotesize\ttfamily,
    keywordstyle=\bfseries,
    commentstyle=\itshape\color{gray},
    breaklines=true,
}

\lstdefinelanguage{BNF}{
    morekeywords={::=, |},
    sensitive=false,
    alsoletter={<>=-},
    moredelim=**[is]{<}{>},
    keywordstyle=\bfseries,
    basicstyle=\ttfamily\footnotesize,
}

%% file: acronym-setup.tex
\usepackage[nolist]{acronym}
\begin{acronym}
\acro{mbse}[MBSE]{\textit{Model-Based Systems Engineering}}
\acro{sysml}[SysML]{\textit{Systems Modeling Language}}
\acro{msosa}[MSoSA]{\textit{Magic Systems of Systems Architect}}
\acro{bnf}[BNF]{\textit{Backus-Naur-Form}}
\acro{pddl}[PDDL]{\textit{Planning Domain Definition Language}}
\acro{uml}[UML]{\textit{Unified Modeling Language}}
\acro{vtl}[VTL]{\textit{Velocity Template Language}}
\acro{ocl}[OCL]{\textit{Object Constraint Language}}
\acro{omg}[OMG]{\textit{Object Management Group}}
\acro{hddl}[HDDL]{\textit{Hierarchical Domain Definition Language}}
\end{acronym}

%% file: figures/workflow.tex
    \begin{tikzpicture}[node distance= 2.9cm]
        \node[exArtifact](sysModel){System Model};
        \node [method, right of= sysModel] (analyzeSM) {Analyze System Model};
        \node [method, right of= analyzeSM] (scope) {Define Scope of Observation};
        \node [method, right of= scope] (identRelElements) {Identify relevant Elements};

        
        \node[exArtifact, below of= sysModel](profile){PDDL Profile};
        \node[method, right of= profile](genDomain){Create PDDL Domain (excl. Actions)};
        \node[method, right of= genDomain](defAction){Define Actions of Domain};
        \node[genArtifact, right of= defAction](aufbSM){Extended System Model};
        \node[below=0.3cm of profile] {\textit{Focus of Section \ref{sec:pddlprofile}}};

        \node[exArtifact, below of= profile](product){Product Model};
        \node[method, right of= aufbSM](genPD){Generate PDDL Descriptions};
        \node[method, right of = product](identProdInfo){Identify Product Information};
        \node[method, right of = identProdInfo](extract){Extract Relevant Information};
        \node[method, right of = extract](transmit){Transfer to MBSE Environment};
        \node[method, right of = transmit](annot){Annotate according to PDDL Domain};
        \node[genArtifact, right of = annot](aufbPM){Extended Product Model};

        \node[exArtifact, above of= genPD](algorithm) {Algorithm};
        \node[genArtifact, right of = genPD](problem){PDDL Domain \& Problem files};
        \node[method, right of = problem](solve){Solve Problem Description};
        \node[genArtifact, above of = solve](plan){PDDL Plan};

        \node[draw, dotted, line width=0.8pt,fit=(sysModel)(analyzeSM)(scope)(identRelElements), inner sep=0.1cm] (phase1) {};
        \node[rotate=90, above=0cm] at (phase1.west) {\textbf{Phase I}};

        \coordinate (boxNW) at ([xshift=-0.09cm,yshift=0.4cm]profile.north west);
        \coordinate (boxSE) at ([xshift=0.08cm,yshift=-0.25cm]aufbSM.south east);
        \node[draw=HighlightPhase, ultra thick, dashed, fit=(boxNW)(boxSE), inner sep=0pt] (phase2) {};
        \node[rotate=90, above=0cm] at (phase2.west) {\textbf{Phase II}};

        \node[draw, dotted, line width=0.8pt,fit=(product)(identProdInfo)(extract)(annot)(aufbPM), inner sep=0.1cm] (phase3) {};
        \node[rotate=90, above=0cm] at (phase3.west) {\textbf{Phase III}};

        \coordinate (boxNW2) at ([xshift=-0.09cm,yshift=0.11cm]algorithm.north west);
        \coordinate (boxSE2) at ([xshift=0.08cm,yshift=-0.25cm]problem.south east);
        \node[draw=HighlightPhase, ultra thick, dashed, fit=(boxNW2)(boxSE2), inner sep=0pt] (phase4) {};
        \node[rotate=90, above=0cm] at (phase2.west) {\textbf{Phase II}};

        \node (phaseIV) at (11.3,1.1) {\textbf{Phase IV}};
        \node[right=0.05cm of phaseIV] {\textit{Focus of Section \ref{sec:algorithm}}};
        \path[line] (sysModel) -- (analyzeSM);
        \path[line] (analyzeSM) -- (scope);
        \path[line] (scope) -- (identRelElements);
        \path[line] (identRelElements) -- ++(0,-1.5cm) -| (genDomain);
        \path[line](profile) -- (genDomain);
        \path[line](genDomain) -- (defAction);
        \path[line](defAction) -- (aufbSM);
        \path[line](aufbSM) -- (genPD);
        \path[line](product) -- (identProdInfo);
        \path[line](identProdInfo) -- (extract);
        \path[line](extract) -- (transmit);
        \path[line](transmit) -- (annot);
        \path[line](annot) -- (aufbPM);
        \path[line](aufbPM) |- ++(0,1.7cm) -| (genPD);
        \path[dashline](genDomain) |- ++(0,-1.5cm)-| (annot);
        \path[line](algorithm)--(genPD);
        \path[line](genPD)  -- (problem);
        \path[line](problem)  -- (solve); 
        \path[line](solve)  -- (plan); 
        
    \end{tikzpicture}